\documentclass[10pt,twocolumn,letterpaper]{article}

\usepackage[pagenumbers]{iccv} 

\usepackage{nicematrix}
\usepackage{dsfont} 
\usepackage{placeins} 
\usepackage{stfloats} 

\usepackage{sg-macros}
\renewcommand{\subsubsection}[1]{\noindent\textbf{#1}}

\newcommand{\partf}{{\mathbf{f}^\text{P}}}
\newcommand{\partfi}{{\mathbf{f}_i^\text{P}}}
\newcommand{\partp}{{\mathbf{p}^\text{P}}}
\newcommand{\stepf}{{\mathbf{f}^\text{I}}}
\newcommand{\stepp}{{\mathbf{p}^\text{I}}}

\definecolor{iccvblue}{rgb}{0.21,0.49,0.74}
\usepackage[pagebackref,breaklinks,colorlinks,allcolors=iccvblue]{hyperref}

\def\paperID{8993} 
\def\confName{ICCV}
\def\confYear{2025}

\title{Manual-PA: Learning 3D Part Assembly from Instruction Diagrams}

\author{
Jiahao Zhang$^{1}$\quad
Anoop Cherian$^{2}$\quad
Cristian Rodriguez$^{3}$\quad
Weijian Deng$^{1}$\quad
Stephen Gould$^{1}$\\
$^1$The Australian National University,
$^2$Mitsubishi Electric Research Labs\\
$^3$The Australian Institute for Machine Learning\\
{\tt\small $^1$\{first.last\}@anu.edu.au}
{\tt\small $^2$cherian@merl.com}
{\tt\small $^3$crodriguezop@gmail.com}
}

\begin{document}
\maketitle


\begin{abstract}
Assembling furniture amounts to solving the discrete-continuous optimization task of selecting the furniture parts to assemble and estimating their connecting poses in a physically realistic manner. The problem is hampered by its combinatorially large yet sparse solution space thus making learning to assemble a challenging task for current machine learning models. In this paper, we attempt to solve this task by leveraging the assembly instructions provided in diagrammatic manuals that typically accompany the furniture parts. Our key insight is to use the cues in these diagrams to split the problem into discrete and continuous phases. Specifically, we present Manual-PA, a transformer-based instruction Manual-guided 3D Part Assembly framework that learns to semantically align 3D parts with their illustrations in the manuals using a contrastive learning backbone towards predicting the assembly order and infers the 6D pose of each part via relating it to the final furniture depicted in the manual. To validate the efficacy of our method, we conduct experiments on the benchmark PartNet dataset. Our results show that using the diagrams and the order of the parts lead to significant improvements in assembly performance against the state of the art. Further, Manual-PA demonstrates strong generalization to real-world IKEA furniture assembly on the IKEA-Manual dataset.
\end{abstract}


\vspace{-10pt}
\section{Introduction}
\label{sec:intro}

\begin{figure}[t]
  \centering
  \includegraphics[width=\linewidth]{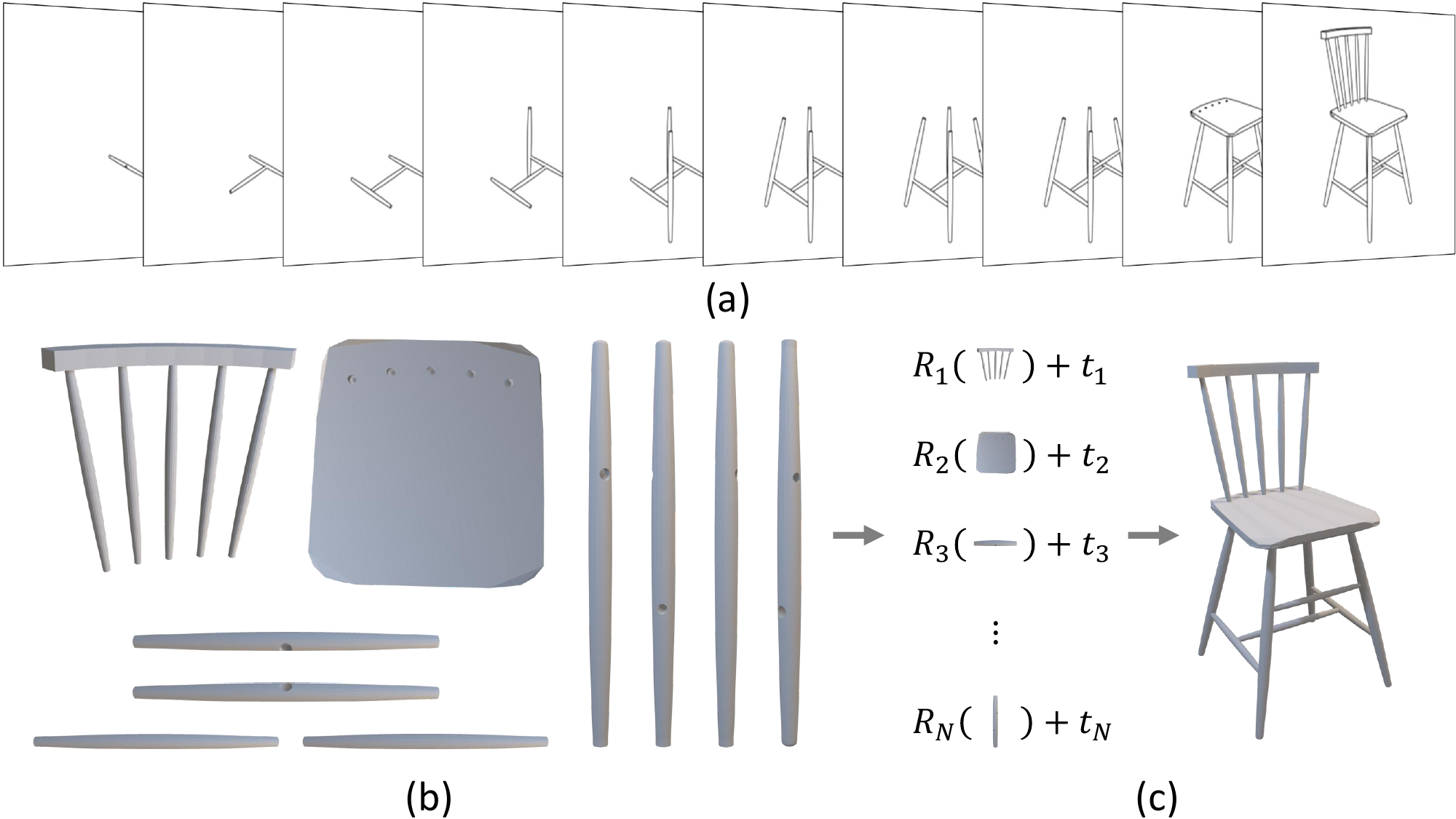}
  \caption{An illustration of the manual-guided 3D part assembly task. Given (a) a diagrammatic manual book demonstrating the step-by-step assembly process and (b) a set of texture-less furniture parts, the goal is to (c) infer the order of parts for the assembly from the manual sequence and predict the 6DoF pose for each part such that the spatially transformed parts assembles the furniture described in the manual.}
  \vspace{-6pt}
  \label{fig:teaser}
\end{figure}

With the rise of the DIY trend, nowadays it is the consumers who are increasingly assembling products like IKEA furniture~\cite{lee2021ikea,ben2021ikea,zhang2023aligning,zhang2024temporally,ben2024ikea}, LEGO structures~\cite{wang2022translating,wan2024scanet}, and exercise equipments~\cite{zhu2023multi,tian2022assemble}. This paradigm shift can reduce the product cost, but often poses challenges for untrained consumers, as an incorrect assembly in any step may lead to irreparable outcomes. Thus, automating the assembly or providing active help in the process is often desired.

Automated assembly faces the dual challenge of large combinatorial solution space, parameterized by the number of discrete parts in the assembly, and the need for a precise 6D pose estimation that correctly connects parts between each other. The limited number of feasible, stable assembly sequences makes this discrete-continuous optimization problem extremely challenging, even for advanced neural networks, especially as the number of parts increases~\cite{xu2024spaformer}. Fortunately, most assembly tasks include language-agnostic instruction manuals, such as diagrams, delineating the assembly. While such manuals help shrink the solution search space, the problem of correctly matching the 2D manual diagrams with their corresponding 3D object parts and inferring their assembly pose and order remains a significant challenge for automated systems.

This paper introduces a method to enable machines to learn how to assemble shapes by following a visual instruction manual, a task we call \textit{manual-guided 3D part assembly}. As illustrated in~\cref{fig:teaser}, given a set of parts and a manual, our approach interprets the manual's step-by-step diagrams to predict the precise 6D pose of each part, assembling them into the target shape. There are two lines of prior approaches that have similarities to our proposed task, namely methods that focus on the geometry of individual parts and inter-part relationships~\cite{li2020learning,zhan2020generative,zhang20223d,cheng2023score,zhang2024scalable,du2024generative}, and methods such as MEPNet~\cite{wang2022translating} that learns the assembly in specific conditions, such as LEGO shapes. The former rely on priors like peg-hole joints~\cite{li2024category} or sequence generation~\cite{xu2024spaformer} to reduce search complexity but are generative and may yield unstable assemblies due to occlusions~\cite{li2020learning}. The latter often simplifies the task by assuming that parts are provided step-by-step, e.g., LEGO manuals. However, furniture manuals, such as IKEA, may not have such information, and it is necessary to determine which parts are used at each step before assembly. Additionally, while LEGO tasks feature standardized ``stud" joints, general furniture assembly is more complex and lacks such constraints.

Based on the above observations, we identify two key challenges in our task. First, how to detect which part or parts are added at each step in the assembly diagrams. Given that these diagrams depict incremental assembly stages, the core challenge is to recognize newly added parts at each step. This requires aligning 2D assembly diagrams with the 3D parts to establish a correct assembly sequence. Second, how to effectively align the assembly process with the learned order. A straightforward approach is to directly use the learned sequence as the order for predicting poses, \eg, by adopting an auto-regressive prediction method similar to MEPNet~\cite{wang2022translating}. However, this approach carries the risk of accumulating errors---if a mistake occurs in an earlier step, subsequent steps are likely to fail as well. On the other hand, completely disregarding the sequence forces the assembly model to implicitly find correspondences, which makes both learning and prediction significantly more difficult. Therefore, the inferred sequence should serve as a \emph{soft guidance} for the assembly process, helping each part focus more on the relevant step diagram.

To address these challenges, we present Manual-guided Part Assembly (Manual-PA), a novel transformer-based neural network that predicts correspondences between step diagrams and parts by computing their semantic similarity, which is learned using contrastive learning. Using this similarity matrix, we establish an assembly order via solving an optimal assignment problem between the aligned features of the 3D object parts and the ordered manual diagram steps, followed by permuting each part's positional encoding according to the learned order. Finally, a transformer decoder fuses the two feature modalities to predict the final 6DoF pose for each part. Notably, the cross-attention between step diagrams and parts is guided by the positional encoding, ensuring that each step diagram receives a higher attention score for its corresponding part.

To empirically validate the effectiveness of Manual-PA, we conduct experiments on the standard PartNet benchmark dataset~\cite{mo2019partnet}, on the categories of assembling the chair, table and storage classes. Our experiments and ablation studies clearly show that Manual-PA leads to state-of-the-art results on multiple metrics, including success rate, outperforming prior methods by a significant margin. Results further showcase the strong generalizability of our approach to real-world IKEA furniture assembly through experiments using the IKEA-Manual dataset~\cite{wang2022ikea}. 

We summarize our key contributions below:
\begin{enumerate}
\item We introduce a new problem setup of assembling 3D parts using diagrammatic manuals for guidance. 

\item We present a novel framework, Manual-PA, for solving this task that learns the assembly order of the 3D parts, which is then used as soft guidance for the assembly.

\item We present experiments on the benchmark PartNet and IKEA-Manual datasets demonstrating state-of-the-art performances and cross-category generalization.
\end{enumerate}


\section{Related Works}
\label{sec:related}

\subsubsection{3D Part Assembly} involves reconstructing complete shapes from sets of 3D parts, a task originally assisted by leveraging semantic part information~\cite{funkhouser2004modeling,sung2017complementme,yin2020coalesce,chaudhuri2011probabilistic,kalogerakis2012probabilistic,wu2020pq}. With the introduction of the PartNet dataset~\cite{mo2019partnet}, recent works~\cite{zhan2020generative,narayan2022rgl,cheng2023score,zhang20223d,du2024generative,zhang2024scalable} leverage geometric features. For instance, 3DHPA~\cite{du2024generative} uses part symmetry for a hierarchical assembly, SPAFormer~\cite{xu2024spaformer} circumvents combinatorial explosion by generating a part order, and Joint-PA~\cite{li2024category} introduces a peg-hole constraint to facilitate assembly. However, these methods are generative towards producing plausible final shapes without aiming for preciseness. Approaches like GPAT~\cite{li2023rearrangement} and Image-PA~\cite{li2020learning} add guidance from point clouds or rendered images, respectively. Our work extends this direction by introducing guidance through instruction manuals, similar to how humans would do.

\subsubsection{Permutation Learning} is often used in self-supervised methods for learning image representations~\cite{noroozi2016unsupervised,santa2017deeppermnet,kim2018learning,carlucci2019domain,pang2020solving}. However, in this work, we seek to align the 3D parts to instruction steps, for which we adopt a contrastive learning ~\cite{kim2021vilt,jia2021scaling,li2021align,radford2021learning} using a novel  cross-modal alignment setup. 

\subsubsection{6DoF Object Pose Estimation} methods can be classified into instance-level~\cite{rad2017bb8,peng2019pvnet,xiang2017posecnn}, category-level~\cite{tian2020shape,wang2019normalized}, and unseen~\cite{labbe2022megapose,wen2024foundationpose}. Our approach relates to the unseen object category as furniture typically comprises novel parts. When the 3D models of these parts are available, many methods~\cite{chen2023zeropose,ornek2023foundpose,nguyen2024gigapose,lin2024sam,huang2024matchu} rely on feature matching followed by the PnP algorithm to infer pose. However, IKEA-style manuals are represented as line drawings making accurate feature extraction for matching to 3D parts challenging. Alternatively, some approaches~\cite{labbe2022megapose,wen2024foundationpose} use end-to-end models to directly predict the 6D pose. A common approach is to use CNOS~\cite{nguyen2023cnos} for object segmentation. However, in assembly diagrams, where multiple parts are connected and depicted together, SAM-based methods~\cite{kirillov2023segment} often struggle with accurate segmentation. Our approach addresses these challenges using an end-to-end framework that uses a cross-attention scheme to enable each part to attend to its relevant region in the step diagram.


\begin{figure*}[t]
  \centering
  \includegraphics[width=\textwidth]{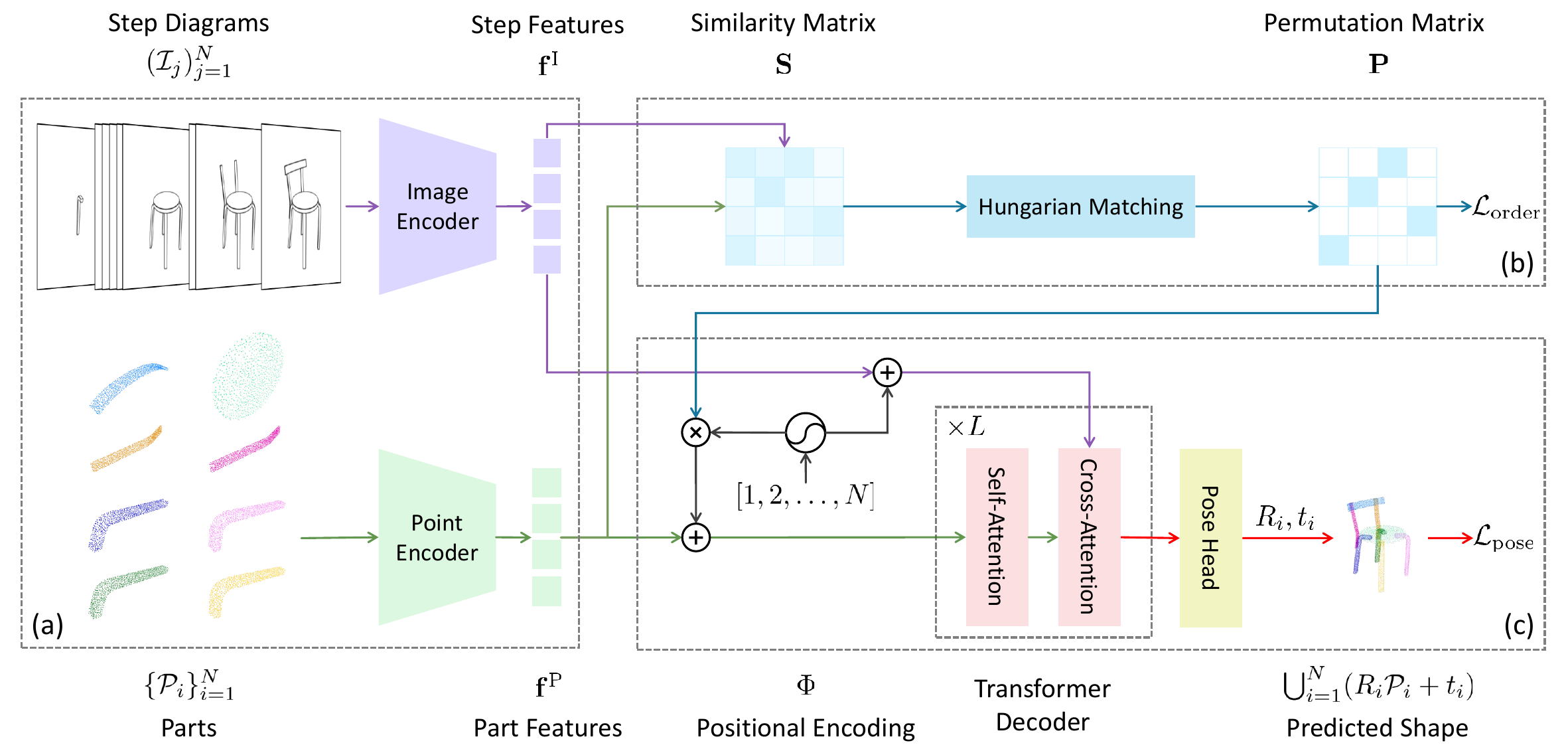}
  \caption{Overview of our proposed method Manual-PA. (a) Feature extraction (\cref{sec:stage0}): we extract semantic and geometrical features from both the step diagrams of the assembly manual and the corresponding part point clouds using the image encoder and point encoder, respectively. (b) Manual-guided part permutation learning (\cref{sec:stage1}): we compute a similarity matrix $\mathbf{S}$ between the two modalities, and subsequently apply the Hungarian algorithm to obtain the permutation matrix $\mathbf{P}$ for the parts. (c) Manual-guided part pose estimation (\cref{sec:stage2}): we add positional encodings (PE) $\Phi$ to both the step diagrams and parts, where the PE order for parts is determined by the order predictions from (b). This is followed by a transformer decoder to enable multimodal feature fusion and interaction, along with a pose prediction head to determine the rotation $R_i$ and translation $t_i$ of each part. The predicted poses are then applied to the corresponding parts to obtain the final assembled shape $\mathbf{S}$.}
  \label{fig:method}
\end{figure*}

\section{Manual-Guided Part Assembly}
\label{sec:method}

In this section, we first provide a concrete formal definition of the \textit{manual-based 3D part assembly} task in~\cref{sec:problem}. We then present a detailed explanation in~\cref{sec:stage0,sec:stage1,sec:stage2} of the proposed learning pipeline as depicted in~\cref{fig:method}, followed by a description of the training process and the loss functions used for each stage in~\cref{sec:losses}.

\subsection{Problem Formulation}
\label{sec:problem}

Our task involves two different input modalities: a set of 3D parts and a corresponding instruction manual. The 3D parts are represented as a multiset\footnote{We allow multiple identical parts, \eg, chair legs, to appear in the set.} of unordered point clouds, denoted by $\{\mathcal{P}_i\}_{i=1}^{N}$. Note that the number of parts $N$ may differ from furniture to furniture. The manual consists of an ordered sequence of $N$ step diagram images, denoted by $(\mathcal{I}_1, \mathcal{I}_2, \ldots, \mathcal{I}_N)$; each diagram depicting an incremental step in the assembly process of a 3D furniture shape $\mathcal{S}$. These step diagrams are 2D line drawings, which are 2D projections of textureless 3D CAD models of these parts captured from a camera $\mathcal{C}$, and we assume for each step, only one additional part is attached to the furniture assembled thus far.  Our goal is to align the 3D parts with their delineations in the respective manual steps towards predicting the assembly order of the parts and the part poses $\{(R_i, t_i)\in SE(3)\}_{i=1}^N$ for the final assembled furniture, where $R_i, t_i$ represent the $3\times 3$ rotation matrix and the three-dimensional translation vector respectively for the $i$-th part in the SE(3) transformation space. That is, these parts undergo rigid transformations such that their combination constructs the complete 3D furniture shape $\mathcal{S}$ as seen from the viewpoint of camera $\mathcal{C}$.

\subsection{Feature Extraction}
\label{sec:stage0}

We first use off-the-shelf encoders to extract semantic latent features from the inputs. For the 3D part point clouds $\{\mathcal{P}_i\}_{i=1}^{N}$, we utilize a lightweight version of PointNet~\cite{qi2017pointnet}, similar to the one used in Image-PA~\cite{li2020learning}, followed by a linear layer to map the features to a common dimension $D$, resulting in geometric part features $\partf\in\reals^{N\times D}$. For sequences of step diagrams $(\mathcal{I}_j)_{j=1}^{N}$, given that the assembly instructions are provided step-by-step in an incremental manner, our primary focus is on the differences between consecutive steps. For any two consecutive diagram images $\mathcal{I}_j$ and $\mathcal{I}_{j+1}$, for $j=1,2,\ldots, N-1$, we compute the difference image given by $|\mathcal{I}_j-\mathcal{I}_{j+1}|$ that shows which new part is added to the thus-far assembled furniture. Next, this difference image is patchified into $K$ image patches, that are then fed to DINOv2 image encoder~\cite{caron2021emerging, oquab2023dinov2} to extract features, followed by a linear layer to map these features to the same dimensionality $D$, resulting in semantic image features $\stepf\in\reals^{N\times K\times D}$.

\subsection{Manual-Guided Part Permutation Learning}
\label{sec:stage1}

In our task, we seek to learn the order of a set of parts conditioned on an ordered sequence of step diagrams. This problem is similar to learning the feature correspondences between two modalities in order to align them, with an ideal alignment corresponding to a permutation matrix. Specifically, our objective is to derive a similarity matrix between the two modalities such that each modality feature is unambiguously assigned to its corresponding feature in the other modality and we desire this assignment matches with the ground truth order in the step sequence in the manual. Formally, we construct the similarity matrix $\mathbf{S}\in\reals^{N\times N}$ as:
\begin{align}
    \mathbf{S}_{ij}=\operatorname{sim}(\partfi, \textbf{g}^\text{I}_j),
\end{align}
where $\textbf{g}^\text{I}\in\reals^{N\times D}$ is derived from the $N\times K\times D$-dimensional step-diagram features $\stepf$ by performing a max-pooling operation on the patch dimension $K$. We measure the similarity in the feature space using the dot product as $\operatorname{sim}(\mathbf{a}, \mathbf{b})=\mathbf{a}\cdot \mathbf{b}$.

Next, we obtain the permutation matrix $\mathbf{P} \in \{0, 1\}^{N \times N}$ by solving a bipartite matching optimization problem using the Hungarian matching algorithm~\cite{kuhn1955hungarian} with cost $\mathbf{C}=-\mathbf{S}$:
\begin{align}
    \renewcommand{\arraystretch}{1.5}
    \begin{array}{rl}
        \text{minimize}   & \sum_{i=1}^N\sum_{j=1}^{N} \mathbf{C}_{ij}\mathbf{P}_{ij} \\
        \text{subject to} & \sum_{j=1}^N P_{ij}=1, \quad \forall i \in \{1, \dots, N\} \\
                          & \sum_{i=1}^N P_{ij}=1, \quad \forall j \in \{1, \dots, N\} \\
                          & P_{ij} \in \{0, 1\}, \quad \forall i, j \in \{1, \dots, N\},
    \end{array}
\label{eq:matching}
\end{align}
where we minimize the cost, subject to constraints ensuring $\mathbf{P}$ is a permutation matrix, i.e., each row and column of $\mathbf{P}$ contains exactly one entry equal to $1$, with all other entries being $0$. The final part order can be obtained by $\sigma = \operatorname{argmax}(\mathbf{P})$ where $\operatorname{argmax}$ is applied columnwise.

\subsection{Manual-Guided Part Pose Estimation}
\label{sec:stage2}

\subsubsection{Positional Encoding.}
Since the attention mechanism in the transformer is permutation-invariant, we follow \citeauthor{vaswani2017attention}~\cite{vaswani2017attention} in using positional encoding to convey the order of both step diagrams and parts to the model. Formally, the positional encoding is defined as a function that maps a scalar to a sinusoidal encoding, $\phi: \mathbb{N} \rightarrow \mathbb{R}^D$:
\begin{align}
    \phi(x)_{2i}=\sin\left(\frac{x}{\tau_p^{2i/d}}\right)\!,\; \phi(x)_{2i-1}=\cos\left(\frac{x}{\tau_p^{2i/d}}\right)\!,
\end{align}
where $i = 1, \dots, d/2$ indexes the dimension and $\tau_p$ is a temperature hyperparameter. By applying $\phi$ to encode the sequential order of step diagrams as they appear in the manual (i.e., using $x$ as a simple increasing sequence $1, 2,\dots, N$), we obtain a positional encoding $\Phi \in \mathbb{R}^{N \times D}$, where $\stepp := \Phi$. Given the permutation matrix $\mathbf{P}$ that corresponds to the order of the parts, we can compute the positional encoding for the parts as $\partp = \mathbf{P}^T\Phi$. During training, we use the ground truth part order, whereas at inference time, we replace it with the predicted permutation matrix $\hat{\mathbf{P}}$, yielding $\hat{\mathbf{p}}^\text{P} = \hat{\mathbf{P}}^T\Phi$. These positional encodings are then added to the respective features before being fed into the transformer decoder and provide soft guidance on the sequence order. We adopt RoPE~\cite{su2024roformer} for better performance.

\subsubsection{Transformer Decoder.}
The modality interaction and fusion between step diagrams and parts occur through a stack of $L$ transformer~\cite{vaswani2017attention} decoder layers. In each layer, the self-attention module receives the output from the previous layer (initially, the part features $\partf + \partp$ in the first layer) to perform part-to-part interactions. This is followed by a cross-attention module that injects and fuses the step diagram features $\stepf + \stepp$ with the part features processed by the self-attention module, enabling step-to-part interactions. Notably, for the step diagram features, we concatenate the features from all diagrams from the same manual along the patch dimension, resulting in a tensor of shape $\reals^{NK \times D}$.

\subsubsection{Pose Prediction Head.}
Finally, the output $O_{\text{Dec}}$ from the transformer decoder is fed into a pose prediction head, which predicts the translation and rotation for each part, denoted as $(\hat{R}_i, \hat{t}_i)$. In our implementation, the rotation matrix $\hat{R}_i$ is represented as a unit quaternion $\hat{q}_i \in \mathbb{R}^4$. We initialize its bias to an identity quaternion $(1, 0, 0, 0)$ and normalize the prediction to ensure a unit norm.

\subsection{Training and Losses}
\label{sec:losses}

First, we train the manual-guided part order learning model until it successfully aligns the two modalities and provides an optimal permutation for the parts. Subsequently, we train the transformer decoder-based part pose estimation model using the predicted order from the first model to infer the part poses. Our empirical results indicate that having an accurate part order is crucial for enhancing the inference performance of the pose estimation network.

\subsubsection{Loss for Permutation Learning.}
Borrowing the notation from~\cref{sec:stage0,sec:stage1}, we define the contrastive-based order loss as: 
\begin{align}
    \mathcal{L}_\text{order}=-\frac{1}{B}\sum_{i=1}^{B}\log\frac{\exp(\operatorname{sim}(\partf_{\!\!\sigma(i)},\textbf{g}^\text{I}_i)/\tau)}{\sum_{j=1}^{B}\exp(\operatorname{sim}(\partf_{\!\!\sigma(i)},\textbf{g}^\text{I}_j)/\tau)},
\end{align}
where $B$ is the mini-batch size, $\tau$ is the temperature, and $\sigma(i)$ denotes the index for the $\sigma(i)$-th part that corresponds to the $j$-th step according to the ground truth part order $\sigma$. Here, mini-batches are constructed by sampling pairs of step-diagram and part features, $(\partf_{\!\!\sigma(i)}, \textbf{g}^\text{I}_j)$ for optimizing the InfoNCE loss~\cite{hadsell2006dimensionality,oord2018representation}. Note that we randomly sample one part from their geometric-equivalent group. In this setup, pairs $(\partf_{\!\!\sigma(i)}, \textbf{g}^\text{I}_i)$ are considered as positive, otherwise negative. This loss encourages positive pairs to have higher similarity while negative pairs have lower similarity, allowing the similarity matrix to be used for generating the permutation matrix via Hungarian matching, as discussed in~\cref{sec:stage1}.

\subsubsection{Losses for Pose Estimation.}
Given a set of furniture parts, there may be subsets of parts that are geometrically identical to each other, such as the four legs of a table, the armrests on both sides of an armchair, or the support rods on the back of a chair. We denote geometrically equivalent parts as a part group $\mathcal{G}$. Within each group, we calculate the chamfer distance between these parts and the ground truth. Then, we apply Hungarian Matching (see~\cref{sec:stage1}) again with cost:
\begin{align}
    \mathbf{C}_{ij}=\operatorname{CD}(\hat{R}_i\mathcal{P}_i+\hat{t}_i, R_j\mathcal{P}_j+t_j),
\end{align}
where CD denotes the chamfer distance measuring the similarity between unordered two point clouds~\cite{fan2017point}, $i$ and $j$ index the parts within $\mathcal{G}$, and $\hat{R}$ and $\hat{t}$ denote the respective predicted rotation and translation. The optimal matching sequence is denoted as $\mathcal{M}$.

For the 3D part assembly task, following \citeauthor{li2020learning}~\cite{li2020learning}, we supervise translation and rotation separately. We compute the $\ell_2$-distance for translation as follows:
\begin{align}
    \mathcal{L}_T=\frac{1}{N}\sum_{i=1}^{N}\lVert \hat{t}_{[i]}-t_i\rVert_2,
\end{align}
where $[i]$ denotes the index of the matched part corresponding to the $i$-th part under optimal matching $\mathcal{M}$.

In addition to the geometric similarity between the parts mentioned above, each part itself may also have intrinsic symmetries. For instance, a cylindrical table leg has axial symmetry. Thus, we cannot simply use the absolute distance (\eg, $\ell_2$) on rotation as a supervision signal. Instead we use chamfer distance to supervise the rotation per part:
\begin{align}
    \mathcal{L}_C=\frac{1}{N}\sum_{i=1}^{N}\operatorname{CD}(\hat{R}_{[i]}\mathcal{P}_i, R_i\mathcal{P}_i).
\end{align}

However, some parts may not be perfectly symmetric, \eg., due to small holes in different locations. To address these cases, we still encourage the model to penalize the $\ell_2$ distance for the rotation as a regularization term, formally expressed as: 
\begin{align}
    \mathcal{L}_E=\frac{1}{N}\sum_{i=1}^{N}\lVert \hat{R}_{[i]}\mathcal{P}_i-R_i\mathcal{P}_i\rVert_2.
\end{align}

Additionally, to ensure that the predicted assembled shape closely matches the ground truth, we compute the chamfer distance between the predicted and ground truth assembled shapes:
\begin{align}
    \mathcal{L}_S=\operatorname{CD}\left(\bigcup_{i=1}^N(\hat{R}_{[i]}\mathcal{P}_i+\hat{t}_{[i]}),\,\bigcup_{i=1}^N(R_i\mathcal{P}_i+t_i)\right),
\end{align}
where $\cup_{i=1}^N$ indicates the union of $N$ parts to form the assembled shape.

Finally, the overall loss for pose estimation is computed as a weighted sum of the aforementioned losses:
\begin{align}
    \mathcal{L}_\text{pose}=\lambda_T\mathcal{L}_T+\lambda_C\mathcal{L}_C+\lambda_E\mathcal{L}_E+\lambda_S\mathcal{L}_S.
\end{align}


\section{Experiments}
\label{sec:experiment}

\subsection{Datasets}

Following~\cite{zhan2020generative,zhang20223d,cheng2023score,zhang2024scalable,du2024generative,narayan2022rgl,xu2024spaformer,li2020learning}, we use PartNet~\cite{mo2019partnet}, a comprehensive 3D shape dataset with fine-grained, hierarchical part segmentation, for both training and evaluation. In our work, we adopt the train/validation/test split used by \citeauthor{li2020learning}~\cite{li2020learning}, where shapes with more than 20 parts are filtered out, and we use the finest granularity, \textit{Level-3}, of PartNet. We focus on the three largest furniture categories: chair, table and storage, containing 40,148, 21,517 and 9,258 parts, respectively. Besides that, to demonstrate the generalizability of our method, we also evaluate its zero-shot capabilities on the IKEA-Manual~\cite{wang2022ikea} dataset, which features real-world IKEA furniture at the part level that are aligned with IKEA manuals. For consistency with PartNet, we evaluate both chair category with 351 parts, and table, with 143 parts. Storage is excluded due to only 3 samples.

For each individual part, we sample 1,000 points from its CAD model and normalize them to a canonical space. Geometrically-equivalent part groups are identified based on the size of their Axis-Aligned Bounding Box (AABB). To generate the assembly manual, we render diagrammatic images of parts using Blender's Freestyle functionality. More details of the preprocessing and rendering process are provided in the supplementary material.

\begin{table*}[t]
    \centering
    \footnotesize
    \newcolumntype{Z}{>{\centering\arraybackslash}X}
    \caption{3D part assembly results on the PartNet test split and Ikea-Manual. \dag{}: We re-trained the Image-PA model using diagrams (2D line drawing images) as the conditioning input instead of the original RGB images. The values in bold represent the best results, while underlined indicate the second. \ddag{}: Average over five runs.}
    \resizebox{\linewidth}{!}{
        \begin{NiceTabularX}{\linewidth}{
                @{}
                l
                c
                |
                *{3}{Z}
                |
                *{3}{Z}
                |
                *{3}{Z}
            }
            \toprule
            \Block{2-1}{Method}
            & \Block{2-1}{Condition}
            & \Block{1-3}{SCD$\downarrow$}
            &
            &
            & \Block{1-3}{PA$\uparrow$}
            &
            &
            & \Block{1-3}{SR$\uparrow$}
            &
            &
            \\\cmidrule(lr){3-5}\cmidrule(lr){6-8}\cmidrule(lr){9-11}
            &
            & Chair
            & Table
            & Storage
            & Chair
            & Table
            & Storage
            & Chair
            & Table
            & Storage
            \\\midrule
            \Block[l]{1-11}{\textit{Fully-Supervised on PartNet~\cite{mo2019partnet}}}
            \\\midrule
            DGL$_\text{NIPS'20}$~\cite{zhan2020generative}
            & -
            & 9.1
            & 5.0
            & -
            & 39.00
            & 49.51
            & -
            & -
            & -
            & -
            \\
            IET$_\text{RA-L'22}$~\cite{zhang20223d}
            & -
            & 5.4
            & 3.5
            & -
            & 62.80
            & 61.67
            &
            & -
            & -
            & -
            \\
            Score-PA$_\text{BMVC'23}$~\cite{cheng2023score}
            & -
            & 7.4
            & 4.5
            & -
            & 42.11
            & 51.55
            & -
            & 8.320
            & 11.23
            & -
            \\
            CCS$_\text{AAAI'24}$~\cite{zhang2024scalable}
            & -
            & 7.0
            & -
            & -
            & 53.59
            & -
            & -
            & -
            & -
            & -
            \\
            3DHPA$_\text{CVPR'24}$~\cite{du2024generative}
            & -
            & 5.1
            & \underline{2.8}
            & -
            & 64.13
            & 64.83
            & -
            & -
            & -
            & -
            \\
            \midrule
            RGL$_\text{WACV'22}$~\cite{narayan2022rgl}
            & Sequence
            & 8.7
            & 4.8
            & -
            & 49.06
            & 54.16
            & -
            & -
            & -
            & -
            \\
            SPAFormer$_\text{ArXiv'24}$~\cite{xu2024spaformer}
            & Sequence
            & 6.7
            & 3.8
            & 4.5
            & 55.88
            & 64.38
            & \underline{56.11}
            & 16.40
            & 33.50
            & \textbf{7.850}
            \\
            Joint-PA$_\text{CVPR'24}$~\cite{li2024category}
            & Joint
            & 6.0
            & 7.0
            & -
            & 72.80
            & 67.40
            & -
            & -
            & -
            & -
            \\
            Image-PA$_\text{ECCV'20}$~\cite{li2020learning}
            & Image
            & 6.7
            & 3.7
            & 5.0
            & 45.40
            & 71.60
            & 40.20
            & -
            & -
            & -
            \\
            Image-PA$_\text{ECCV'20}^\dag{}$
            & Diagram
            & 5.9
            & 3.9
            & 3.7
            & 62.67
            & 70.10
            & 47.79
            & 19.97
            & 32.83
            & 4.730
            \\
            \midrule
            Manual-PA w/o Order
            & Manual
            & \underline{3.0}
            & 3.6
            & 4.2
            & 79.07
            & 74.03
            & 48.54
            & 34.13
            & 37.71
            & 4.054
            \\
            Manual-PA (Ours)
            & Manual
            & \textbf{1.7}
            & \textbf{1.8}
            & \textbf{2.9}
            & \textbf{89.06}
            & \textbf{87.41}
            & \textbf{56.51}
            & \textbf{58.03}
            & \textbf{56.66}
            & 4.730
            \\
            \midrule
            Manual-PA (Ours)
            & Multi-Part
            & 3.4
            & 3.5
            & 3.3
            & \underline{84.63}
            & \underline{78.95}
            & 53.41
            & \underline{38.31}
            & \underline{40.15}
            & \underline{5.405}
            \\
            Manual-PA (Ours)
            & Multi-View$^\ddag{}$
            & 3.6
            & 3.2
            & \underline{3.0}
            & 82.04
            & 75.49
            & 54.37
            & \underline{38.31}
            & 39.96
            & 4.054
            \\\midrule
            \Block[l]{1-11}{\textit{Zero-Shot on IKEA-Manual~\cite{wang2022ikea}}}
            \\\midrule
            3DHPA$_\text{CVPR'24}$
            & -
            & 34.3
            & 37.8
            & -
            & 1.914
            & 4.027
            & -
            & 0.000
            & 0.000
            & -
            \\
            Image-PA$_\text{ECCV'20}^\dag{}$
            & Diagram
            & 17.3
            & 14.7
            & -
            & 19.07
            & 36.74
            & -
            & 0.000
            & \underline{10.53}
            & -
            \\
            \midrule
            Manual-PA w/o Order
            & Manual
            & \underline{12.8}
            & \underline{8.9}
            & -
            & \underline{38.36}
            & \underline{42.01}
            & -
            & \underline{1.754}
            & \underline{10.53}
            & -
            \\
            Manual-PA (Ours)
            & Manual
            & \textbf{11.4}
            & \textbf{4.8}
            & -
            & \textbf{42.51}
            & \textbf{49.72}
            & -
            & \textbf{3.509}
            & \textbf{15.79}
            & -
            \\
            \bottomrule
        \end{NiceTabularX}
    }
    \label{tbl:assembly_sota}
\end{table*}

\subsection{Evaluation Metrics}

To evaluate the effectiveness of 3D part assembly of different models, we employ three key metrics: Shape Chamfer Distance (SCD)~\cite{li2020learning}, Part Accuracy (PA)~\cite{li2020learning}, and Success Rate (SR)~\cite{li2023rearrangement}. SCD quantifies the overall chamfer distance between predicted and ground truth shapes, providing a direct measure of assembly quality, scaled by a factor of $10^3$ for interpretability. PA assesses the correctness of individual part poses by determining whether the chamfer distance for a part falls below a threshold of $0.01$, reflecting the accuracy of each part's alignment. Finally, SR evaluates whether all parts in an assembled shape are accurately predicted, offering a strict criterion for complete assembly success. 

\subsection{Main Results}

\subsubsection{Fully Supervised on PartNet.}
In this setting, we train our model on PartNet for each category and test it on unseen shapes in the test split, which share similar patterns with the training set. As shown in \cref{tbl:assembly_sota}, our method, Manual-PA, significantly outperforms all previous approaches in all three categories. Specifically, Manual-PA achieves superior performance compared to all other conditioned methods. Notably, when compared to Image-PA, we observe improvements of 26, 17 and 8 part accuracy for the chair, table and storage categories, respectively. This shows that for the 3D Part Assembly task, instruction manuals provide far better guidance than a single image or other conditioning inputs. ``Manual-PA w/o Order'' refers to our model without access to the part assembly order information, requiring it to learn this implicitly. Although the instruction manual provides strong visual guidance, performance improvements remain limited in this case, highlighting the challenges of implicit learning. By explicitly learning the correspondences between step diagrams and parts and applying positional encoding as soft guidance, we achieve PA improvements of approximately 10, 13 and 8 in the chair, table and storage categories, respectively. Note that all other methods, except for Image-PA and ours, are generative and follow the setting proposed by \cite{zhan2020generative}, where multiple assembly results are generated with varying Gaussian noise for the same set of parts. Only the most faithful assembly (based on Minimum Matching Distance~\cite{achlioptas2018learning}) is evaluated.

\subsubsection{Extension to More Realistic Assembly Settings.}
Our problem setting arises from the lack of a standardized format in real-world instruction manuals, which makes it difficult to define a universal learning framework. Our model is designed with scalability and generalizability in mind, allowing it to be extended to more realistic assembly scenarios. To demonstrate this, we implement and evaluate our framework under two challenging settings: 1) Multi-Part Assembly, where multiple new parts are introduced in a single step. In this scenario, parts are grouped based on their geometric similarity, and their positional encodings are adjusted accordingly to reflect their associations. 2) Multi-View , where each step diagram is captured from a randomly chosen viewpoint among eight predefined angles, simulating the viewpoint variations commonly observed in real manuals. To handle these variations, we modify the vision encoder to encode consecutive step diagrams separately before concatenating their features as input. As shown in~\cref{tbl:assembly_sota}, while these more complex settings introduce additional challenges and result in some performance degradation compared to the original setup, our method still significantly outperforms baseline approaches. These results highlight the robustness and versatility of our framework, demonstrating its ability to generalize beyond the single-part-per-step assumption and handle more practical assembly tasks. Further details on these experiments are provided in the supplementary material.

\begin{figure}[t]
  \centering
  \includegraphics[width=\linewidth]{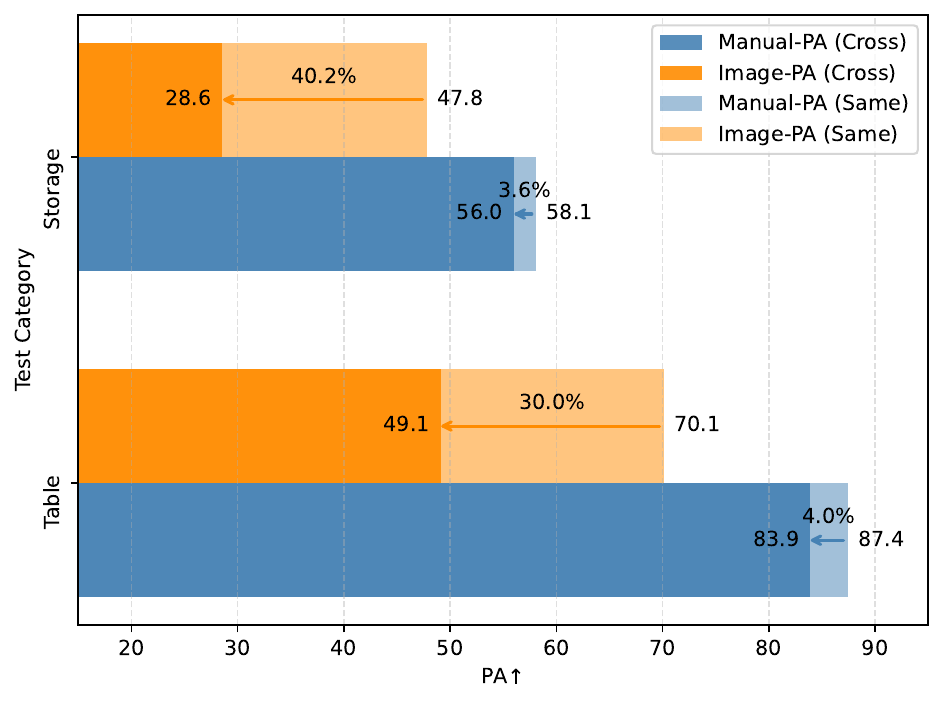}
  \caption{Results trained on chair and tested table and storage. Solid bars: cross-category; lighter bars: same-category. Arrows show performance drop with percentages above.}
  \label{fig:cross}
\end{figure}

\subsubsection{Cross-Category Generalization.}
We evaluate cross-category generalization by training both Manual-PA and Image-PA on the chair category and testing on chair, table, and storage. As shown in~\cref{fig:cross}, the performance drop of Manual-PA when tested on unseen categories is minimal compared to that of Image-PA, indicating stronger cross-category generalization. This result suggests that Manual-PA does not solely rely on shape priors but effectively extracts assembly-relevant information from manuals, enabling strong performance even on unseen shapes. These findings reinforce the advantage of using stepwise instruction diagrams as guidance for 3D part assembly.

\subsubsection{Zero-shot on IKEA-Manual.}
We use the same model, pretrained on PartNet, to directly estimate the poses of furniture in IKEA-Manual, where all shapes are 3D models of real-world IKEA furniture, and each part corresponds to illustrations in their official assembly manuals. As shown in~\cref{tbl:assembly_sota}, our method, significantly outperforms strong competitors such as 3DHPA and Image-PA, demonstrating the strong generalizability of our approach to real world.

\subsection{Ablation Studies}

\begin{table}[t]
    \centering
    \footnotesize
    \newcolumntype{Z}{>{\centering\arraybackslash}X}
    \caption{Ablation results for various order designs on the PartNet chair test split. We use GT order for both training and inference.}
    \resizebox{\linewidth}{!}{
        \begin{NiceTabularX}{\linewidth}{
                @{}
                l
                l
                c
                c
                |
                *{3}{Z}
            }
            \toprule
            \#
            & Type
            & Step
            & Parts
            & SCD$\downarrow$
            & PA$\uparrow$
            & SR$\uparrow$
            \\\midrule
            1
            & PE
            &
            &
            & 4.3
            & 72.22
            & 22.50
            \\
            2
            & PE
            &
            &
            \checkmark
            & 3.0
            & 79.31
            & 33.88
            \\
            3
            & OneHot
            &
            &
            \checkmark
            & 3.0
            & 79.07
            & 34.13
            \\
            4
            & OneHot
            &
            \checkmark
            &
            \checkmark
            & 2.7
            & 81.22
            & 36.92
            \\
            5
            & PE (Ours)
            &
            \checkmark
            &
            \checkmark
            & 1.7
            & 95.38
            & 73.07
            \\
            \bottomrule
        \end{NiceTabularX}
    }
    \label{tbl:pe_ablation}
\end{table}

\begin{table}[t]
    \centering
    \footnotesize
    \newcolumntype{Z}{>{\centering\arraybackslash}X}
    \caption{Ablation results for various component designs in permutation learning on the PartNet chair test split. KT denotes the Kendall-tau coefficient that measure the ordinal correlation between predicted and GT order.}
    \resizebox{\linewidth}{!}{
        \begin{NiceTabularX}{\linewidth}{
                @{}
                l
                |
                *{4}{Z}
            }
            \toprule
            Variants
            & KT$\uparrow$
            & SCD$\downarrow$
            & PA$\uparrow$
            & SR$\uparrow$
            \\\midrule
            Manual-PA (Ours)
            & 0.789
            & 1.7
            & 89.06
            & 58.03
            \\
            \quad w/o batch-level sampling
            & 0.410
            & 2.3
            & 56.93
            & 15.68
            \\
            \quad w/ Sinkhorn
            & 0.774
            & 1.7
            & 88.25
            & 56.26
            \\
            \quad w/ GT Order
            & 1.000
            & 1.7
            & 95.38
            & 73.07
            \\
            \bottomrule
        \end{NiceTabularX}
    }
    \label{tbl:order_ablation}
\end{table}

\subsubsection{Effect of the Order for 3D Part Assembly.}
We conduct experiments to explore various methods for incorporating order information in a 3D part assembly model, as shown in~\cref{tbl:pe_ablation}. First, by comparing lines 1 and 2, we observe that adding sequential information solely to part features improves assembly performance. We attribute this improvement to two factors: (1) distinguishing geometrically equivalent components, and (2) mitigating the combinatorial explosion problem, as claimed in SPAFormer~\cite{xu2024spaformer}. Second, by comparing lines 2 and 5, where the latter provides both the step diagram and part features with correct correspondence, we achieve a significant performance boost. This indicates that explicitly establishing correspondence between steps and parts is beneficial, as learning these correspondences implicitly remains challenging for the model. Last, our empirical results show that, when position information is solely relevant to self-attention (lines 2 and 3), both PE and OneHot encodings yield similar performance. This observation aligns with prior transformer-based studies, which also utilize OneHot encodings~\cite{zhang20223d,du2024generative,xu2024spaformer}. However, in cases requiring cross-attention (line 4 and 5), PE outperforms OneHot encodings, highlighting its advantages in cross-attentive contexts.

\subsubsection{Analysis of Permutation Learning Choices.}
We conduct an ablation study for permutation learning, as shown in~\cref{tbl:order_ablation}. First, we find that batch-level sampling is essential compared to applying contrastive learning individually within each manual book. This is primarily because contrastive learning benefits from a larger pool of negative samples. Second, while we employ the Sinkhorn~\cite{sinkhorn1967diagonal} algorithm on the similarity matrix to enforce that rows and columns sum to one, empirical results indicate limited performance improvement. Last, by providing the model with the ground truth assembly order, we establish an upper performance bound for our assembly model.

\begin{figure*}[t]
  \centering
  \includegraphics[width=\linewidth]{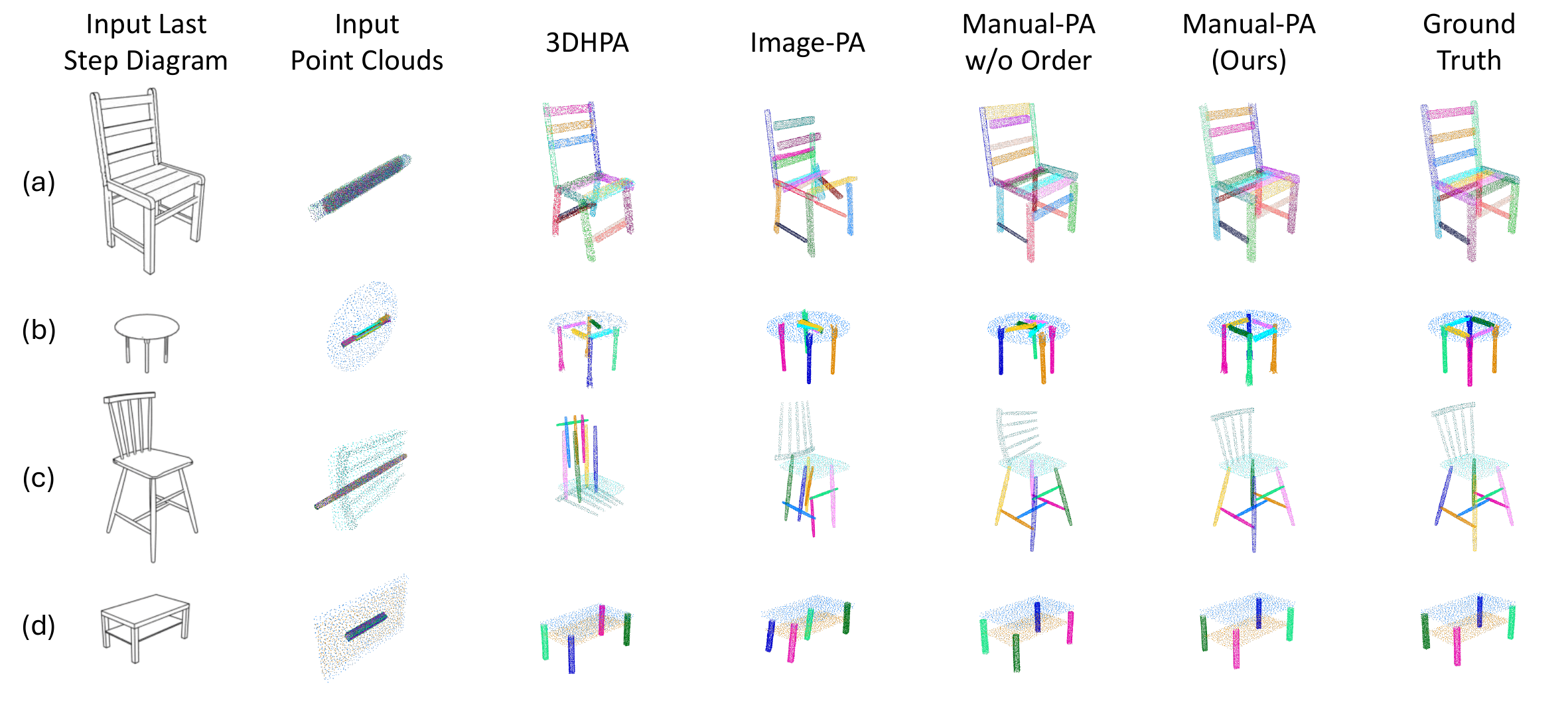}
  \caption{Qualitative comparison of various 3D part assembly methods. Four examples are shown: chair (a) and table (b) from the PartNet dataset, and chair (c) and table (d) from the IKEA-Manual dataset.}
  \label{fig:vis}
\end{figure*}

\begin{figure}[t]
  \centering
  \includegraphics[width=\linewidth]{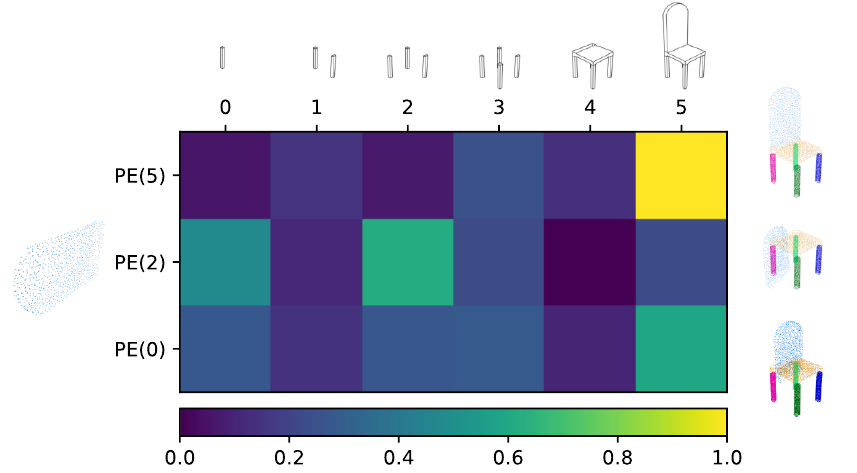}
  \caption{Visualization of the cross attention scores between step diagrams (top) and a part (left, chair's back) with PE at different positions. The final assembly results are displayed on the right.}
  \label{fig:attention}
\end{figure}

\subsection{Visualization and Analysis}

\subsubsection{Demonstration of Soft Guidance.} 
We manually add different  positional encodings to the feature of chair's back, as shown in~\cref{fig:attention}. With PE(5), the model focuses on the final position, aligning well with the correct step diagram. However, when PE(2) is applied, drawing the model’s attention to the second step diagram, the model struggles to accurately predict the chair's back pose. Notably, even with PE(0), the model manages to shift its attention toward the correct final step, demonstrating the fault tolerance and robustness of soft guidance: the model can still identify the correct match even when given an incorrect sequence.

\subsubsection{Qualitative Comparison.}
As shown in~\cref{fig:vis}, our proposed Manual-PA produces the most faithful part pose prediction for the shapes illustrated in the step diagrams. For ``Manual-PA w/o Order'', some parts do not align well with the assembly instructions, indicating the importance of providing precise correspondences between the step diagram and the parts for accurate assembly. In the case of Image-PA, the input consists only of the final image, making it challenging to locate the poses of parts that are occluded, such as the supporting rods beneath the table top in (b). 3DHPA lacks visual input and operates in a generative mode, relying solely on shape priors. Consequently, it generates shapes that are reasonable, as seen in (a), but there are noticeable discrepancies between its output and the ground truth shapes. For zero-shot results on the IKEA-Manual dataset, both 3DHPA and Image-PA exhibit degraded performance, while Manual-PA still produces faithful assembly results, demonstrating its superior generalization ability.


\section{Conclusions and Future Work}
\label{sec:conclusion}

We introduce a novel 3D part assembly framework that leverages diagrammatic manuals for guidance. Our proposed approach, Manual-PA, learns the sequential order of part assembly and uses it as a form of soft guidance for 3D part assembly. Manual-PA achieves superior performance on the PartNet dataset compared to existing methods and demonstrates strong generalizability to real-world IKEA furniture, as validated on the IKEA-Manual dataset.

Future work could aim to bridge the gap between synthetic and real-world IKEA manuals. Although we have addressed adaptability in detecting a varying number of new parts introduced in each step and robust handling of differing perspectives, other visual elements such as arrows, detailed close-ups, and part labels remain challenging. Moreover, manuals often include sub-module assemblies, resulting in a tree-like, non-linear assembly structure.

{
    \small
    \bibliographystyle{ieeenat_fullname}
    \bibliography{main}
}

\end{document}


\maketitle
\appendix

\tableofcontents

\section{Implementation Details}

We train our model using AdamW~\cite{loshchilov2018decoupled} with a learning rate of $10^{-5}$ and a weight decay of $10^{-4}$. For permutation learning, the learning rate decays by a factor of $0.9$ every 5 epochs across a total of 50 epochs, whereas for pose estimation, it decays by the same factor every 50 epochs over 1000 epochs. All experiments are conducted on a single A100 GPU with 80GB of memory, requiring approximately 70 hours for the PartNet chair category, 40 hours for the table category and 16 hours for the storage category. Following~\cite{li2020learning}, we set the hyperparameters as $\lambda_T = 1$, $\lambda_E = 1$, $\lambda_C = 20$, and $\lambda_S = 20$. The code and dataset will be made publicly available upon acceptance.

\section{Metric Details}

\subsubsection{Shape Chamfer Distance (SCD)}~\cite{li2020learning} provides a direct measure of the overall chamfer distance between predicted and ground truth shapes. Using notations introduced in~Secs. 3.4 and 3.5, SCD is defined as:
\begin{align}
    \operatorname{SCD}=\operatorname{CD}\left(\bigcup_{i=1}^N(\hat{R}_{[i]}\mathcal{P}_i+\hat{t}_{[i]}),\bigcup_{i=1}^N(R_i\mathcal{P}_i+t_i)\right),
\end{align}
where $\cup_{i=1}^N$ indicates the union of $N$ parts to form the assembled shape and $[i]$ denotes the index of the matched part corresponding to the $i$-th part under optimal matching $\mathcal{M}$. SCD is scaled by a factor of $10^3$ for better practical interpretation.

\subsubsection{Part Accuracy (PA)}~\cite{li2020learning} assesses the correctness for individual part poses. 
A part pose is considered as accurate if its chamfer distance is below a specific threshold $\epsilon=0.01$:
\begin{align}
    \operatorname{PA}=\frac{1}{N}\sum_{i=1}^N\mathds{1}\left(\operatorname{CD}(\hat{R}_{[i]}\mathcal{P}_i+\hat{t}_{[i]}, R_i\mathcal{P}_i+t_i)<\epsilon\right),
\end{align}
where $\mathds{1}$ is an indicator function, which returns 1 if the condition is met and 0 otherwise.

\subsubsection{Success Rate (SR)}~\cite{li2023rearrangement} is $1$ if all parts in a shape are considered as accurate, and $0$ otherwise:
\begin{align}
    \operatorname{SR}=\mathds{1}(\operatorname{PA}=1).
\end{align}

This metric measures whether an entire assembled shape is correctly predicted, making it a stringent criterion for evaluating complete assembly success.

\subsubsection{Kendall-Tau (KT)}~\cite{santa2017deeppermnet} measures the ordinal correlation between the ground truth permutation $\sigma \in \candinals^N$ and the predicted permutation $\hat{\sigma} \in \candinals^N$. Formally, KT is defined as:
\begin{align}
    \operatorname{KT}=\frac{c^+(\hat{\sigma},\sigma)-c^-(\hat{\sigma},\sigma)}{N(N-1)/2},
\end{align}
where $c^+(\hat{\sigma}, \sigma)$ and $c^-(\hat{\sigma}, \sigma)$ denote the number of correctly (concordant) and incorrectly (discordant) ordered pairs in the sequence, respectively. The KT metric ranges from $-1$ to $1$, where $1$ indicates a perfect match, $-1$ indicates a completely reversed sequence, and $0$ represents a random ordering.

\section{Dataset Creation Details}

\begin{figure*}[ht]
  \centering
  \vspace{.4cm}
  \includegraphics[width=\linewidth]{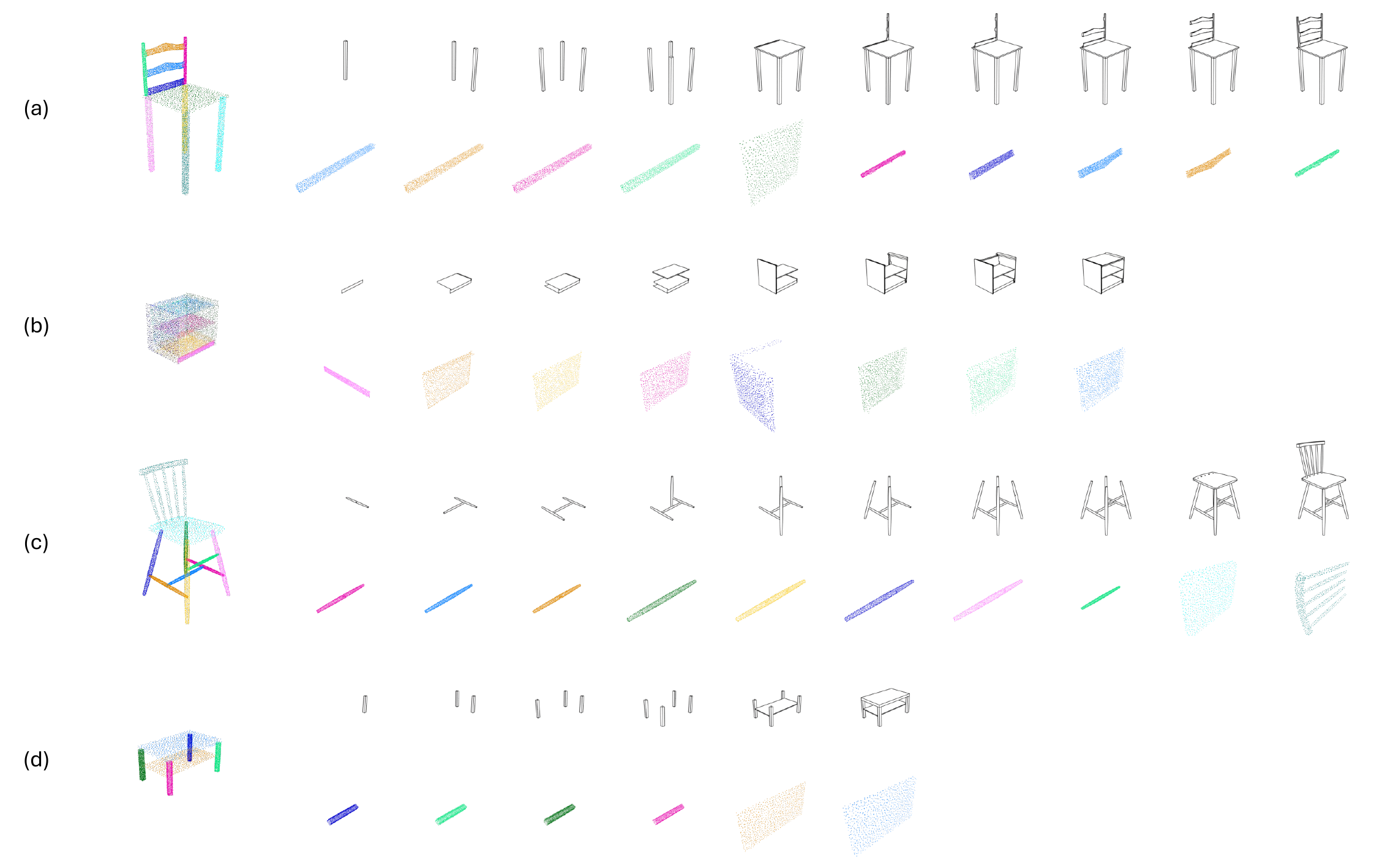}
  \vspace{.3cm}
  \caption{Illustration of several examples from the datasets. (a) and (b) are from the PartNet dataset, while (c) and (d) are from the IKEA-Manual dataset. (a) and (c) are examples of chairs, and (b) and (d) are examples of tables. For each example, the leftmost column shows the fully assembled shape in their point cloud format, the top row on the right presents the step-by-step assembly instruction manual we generated, and the bottom row displays the point cloud of the newly added parts for each step.}
  \label{fig:manual}
\end{figure*}

We reuse the dataset provided by~\citeauthor{li2020learning}~\cite{li2020learning} for PartNet and apply the same preprocessing pipeline for IKEA-Manual. Specifically, for each part, we first randomly select 10,000 vertices from its mesh model, followed by sampling 1,000 points using Farthest Point Sampling (FPS). All point clouds are normalized to be centered at the world origin, adopting a canonical coordinate system derived via Principal Component Analysis (PCA)~\cite{pearson1901liii}. The longest diagonal of their Axis-Aligned Bounding Box (AABB) is scaled to unit length, eliminating the effects of scale variations across furniture. Additionally, to group geometrically similar parts, we adopt a robust heuristic based on AABB diagonal lengths. Parts with similar diagonal lengths are classified into the same group, regardless of their original pose or orientation. This grouping accounts for symmetries and ensures that similar components, such as table legs or chair arms, are treated uniformly during the assembly process.

As shown in~\cref{fig:manual}, for both the PartNet and IKEA-Manual datasets, we generate step-by-step furniture assembly manuals to simulate real-world instructional guides. Using Blender's~\cite{blender,denninger2023blenderproc2} Freestyle functionality, we render 2D line drawing diagrams. First, the fully assembled furniture is placed at the world origin, with the camera positioned to provide a clear frontal view. For each subsequent assembly step, one part is removed, and the scene is re-rendered. Freestyle’s edge-enhancement capabilities ensure that the diagrams highlight the part edges effectively, resembling traditional technical manuals. The assembly order is determined using two complementary criteria: (1) parts grouped by AABB diagonal lengths are ordered from bottom to top along the z-axis, reflecting a natural bottom-up assembly process, and (2) within each group, parts are ordered by their distance from the camera, starting with the farthest. This ensures visibility and interpretability of the remaining parts in the diagram.

\clearpage

\section{More Experiment Results}

\subsection{Details of Multi-Step and Multi-View.}

To extend our framework to more realistic assembly scenarios, we implement two modifications: Multi-Part, which allows multiple parts to be added in a single step, and Multi-View, which introduces viewpoint variations across steps. Both modifications are performed by fine-tuning our previously trained model, as it already possesses the foundational ability for manual-guided 3D part assembly.

\subsubsection{Multi-Step.}
Motivated by the observation that real-world instruction manuals frequently introduce multiple parts simultaneously, we extend our framework to handle Multi-Part scenarios. This allows us to assess our model's capability in more complex and realistic assembly tasks. Specifically, given an initial set of $N$ parts, we group these parts into $M$ groups (where $M\leq $) based on geometric similarity computed using Chamfer Distance. Each group of similar parts is then associated with a single step diagram. Consequently, we adjust the positional encodings according to these $M$ groups rather than individual parts. Importantly, this grouping simplifies positional encoding assignments, and the model's input is updated accordingly to reflect these $M$ groups. The similarity matrix used within our framework thus becomes an $M\times M$ square matrix

\subsubsection{Multi-View}
Real-world manuals frequently present steps from varying viewpoints to avoid occlusions and clearly illustrate part connections. To reflect this practical scenario, we define a set of eight predefined viewpoints, corresponding to the vertices of a 3D cube surrounding the object. During dataset preparation, each step diagram's viewpoint is randomly selected from these eight viewpoints, introducing realistic visual variation across assembly steps. To facilitate viewpoint alignment, we initialize each instruction manual with an additional blank diagram as the first step, ensuring consistent reference for subsequent viewpoint encoding. Moreover, to maintain canonical space across training, we enforce that the final step diagram always uses a fixed viewpoint (denoted as View 0). In our model, consecutive step diagrams, including the blank initial step, are individually processed through the vision encoder, yielding feature representations $f_t$ and $f_{t+1}$. These features are then concatenated into $f'=[f_t;f_{t+1}]$, allowing the model to effectively capture the relationships and viewpoint changes between adjacent steps, thereby enhancing robustness to varying visual perspectives.

\subsection{Number of Parts.}
\begin{figure}[t]
  \centering
  \includegraphics[width=\linewidth]{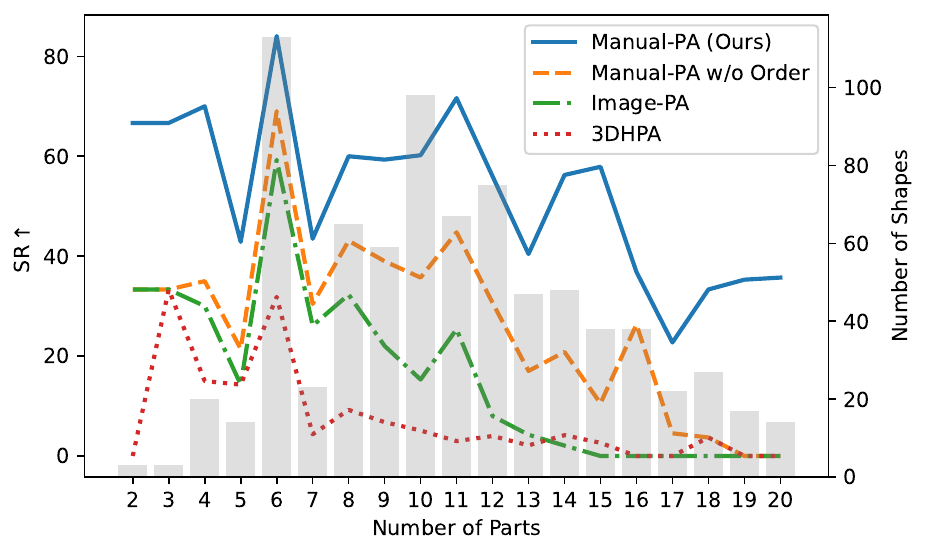}
  \caption{Comparison of average success rates (SR) across varying numbers of parts for different methods on PartNet chair test split. Number of chairs tested shown in background bar chart.}
  \label{fig:nparts}
\end{figure}
We analyze the assembly quality across varying numbers of parts on PartNet chair test split, as shown in~\cref{fig:nparts}. It is generally accepted that, assembling objects with more parts is more challenging due to the combinatorial explosive issue. However, our method consistently achieves a higher success rate across all part counts, demonstrating robust adaptability to different levels of assembly complexity. Notably, for shapes with more than 10 parts, 3DHPA and Image-PA exhibit near-zero success rates, whereas our method, Manual-PA, continues to produce competitive assembly results.

\subsection{More Ablation Studies}
\begin{table}[t]
    \centering
    \footnotesize
    \newcolumntype{Z}{>{\centering\arraybackslash}X}
    \caption{3D part assembly results on the PartNet chair test split. We train and test the model with ground truth order.}
    \vspace{.3cm}
    \resizebox{\linewidth}{!}{
        \begin{NiceTabularX}{\linewidth}{
                @{}
                l
                |
                *{3}{Z}
            }
            \toprule
            Exp.
            & SCD$\downarrow$
            & PA$\uparrow$
            & SR$\uparrow$
            \\\midrule
            Manual-PA (Ours)
            & \textbf{1.7}
            & \textbf{95.38}
            & \textbf{73.07}
            \\
            \quad w/o RoPE
            & \underline{1.8}
            & \underline{94.80}
            & \underline{69.53}
            \\
            \quad w/ 3 Decoder Layers
            & 2.8
            & 86.63
            & 42.73
            \\
            \bottomrule
        \end{NiceTabularX}
    }
    \label{tbl:more_ablation}
\end{table}
As shown in~\cref{tbl:more_ablation}, we conduct two additional ablation studies. In the first study, we incorporate RoPE (Rotary Position Embedding)~\cite{su2024roformer} into the attention mechanism. We observe that RoPE does not conflict with the pre-existing positional encoding (PE) in the features and that its inclusion further improves the model's performance. In the second study, we examine the impact of the number of transformer decoder layers. In our default model, the decoder consists of six layers. When we reduce the number of layers to three, the performance drops significantly, particularly in the SR metric, which decreases by 27 percentage points. These results highlight the importance of using more decoder layers for the 3D part assembly task.

\subsection{Kendall Tau vs. Performance}

\begin{figure}[t]
  \centering
  \includegraphics[width=\linewidth]{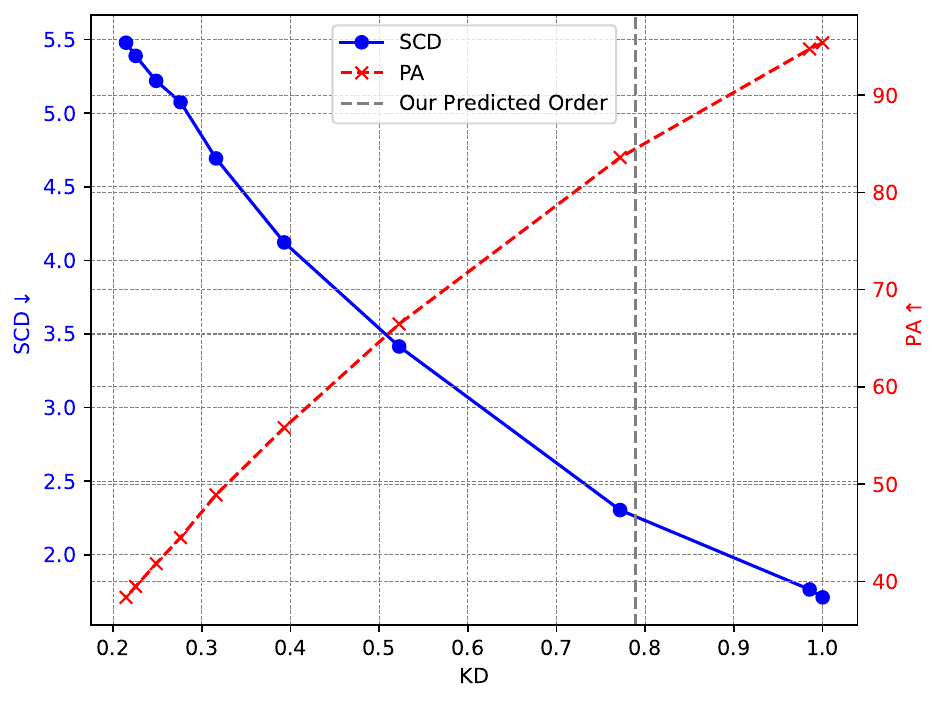}
  \caption{The impact of different part orders on the performance of the 3D part assembly model evaluated on PartNet chair test split. As the Kendall Tau (KD) approaches 1, the order of the parts becomes increasingly correlated with the order specified in the manual, whereas lower KD values indicate less correlation.}
  \label{fig:order}
\end{figure}

As shown in~\cref{fig:order}, we conduct experiments to investigate the impact of part order on the performance of the 3D part assembly model. Starting with a model trained using the ground truth order, we introduce varying levels of Gumbel noise to the permutation matrix to randomly perturb the order and then perform inference for 3D part assembly. The results reveal that assembly performance improves as Kendall Tau (KD) approaches 1, indicating a stronger correlation between the perturbed and ground truth orders. Conversely, lower KD values lead to poorer performance, which aligns with intuition: incorrect correspondences make it difficult for the model to identify the correct step diagram for each part, thereby hindering pose prediction. The results also highlight the performance upper bound of our method when $\operatorname{KD} = 1$. Interestingly, the order learned through permutation learning achieves a KD of approximately 0.79, outperforming randomly perturbed orders with similar KD values. This advantage stems from the fact that our method's randomness primarily arises from the indeterminacy of part order within geometrically equivalent groups, while maintaining relatively accurate alignment across groups.

\subsection{Cross Attention Map on Step Diagrams.}
\begin{figure}[t]
  \centering
  \includegraphics[width=\linewidth]{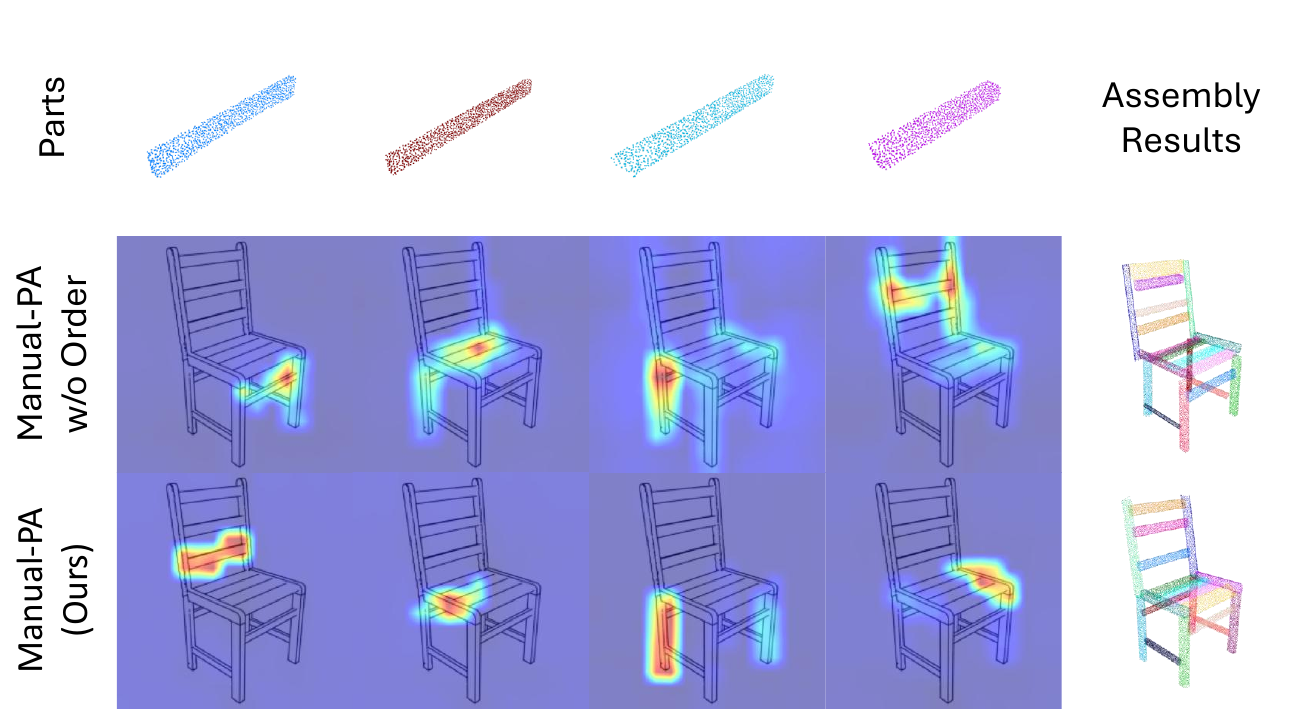}
  \caption{Visualization of cross-attention maps on step diagrams. The cross-attention represents the mean aggregation for each part across all step diagrams. Color red indicates a higher attention.}
  \label{fig:diagram_attention}
\end{figure}
As shown in~\cref{fig:diagram_attention}, we visualize the cross-attention maps between each part and the step diagrams. High attention values correspond to regions in the step diagram where the model pays more attention, which aligns with the spatial placement of parts during assembly. For instance, when the first blue board focuses on the back of the chair, it is positioned on the chair back, and when it attends to the area under the seat, it acts as a support between the legs. Comparing the two methods, we observe that without order as a soft guidance (represented by ``Manual-PA w/o Order''), the attention regions are more dispersed, resulting in less accurate part placement. For example, the third cyan board does not focus entirely on the left chair leg without soft guidance, leading to a misaligned leg position. The absence of order as explicit guidance also results in incorrect part placement. For instance, the fourth purple board, which should be part of the seat, is instead assembled onto the chair back.

\clearpage

\section{More Qualitative Results}
\begin{figure*}[b]
  \centering
  \includegraphics[width=\linewidth]{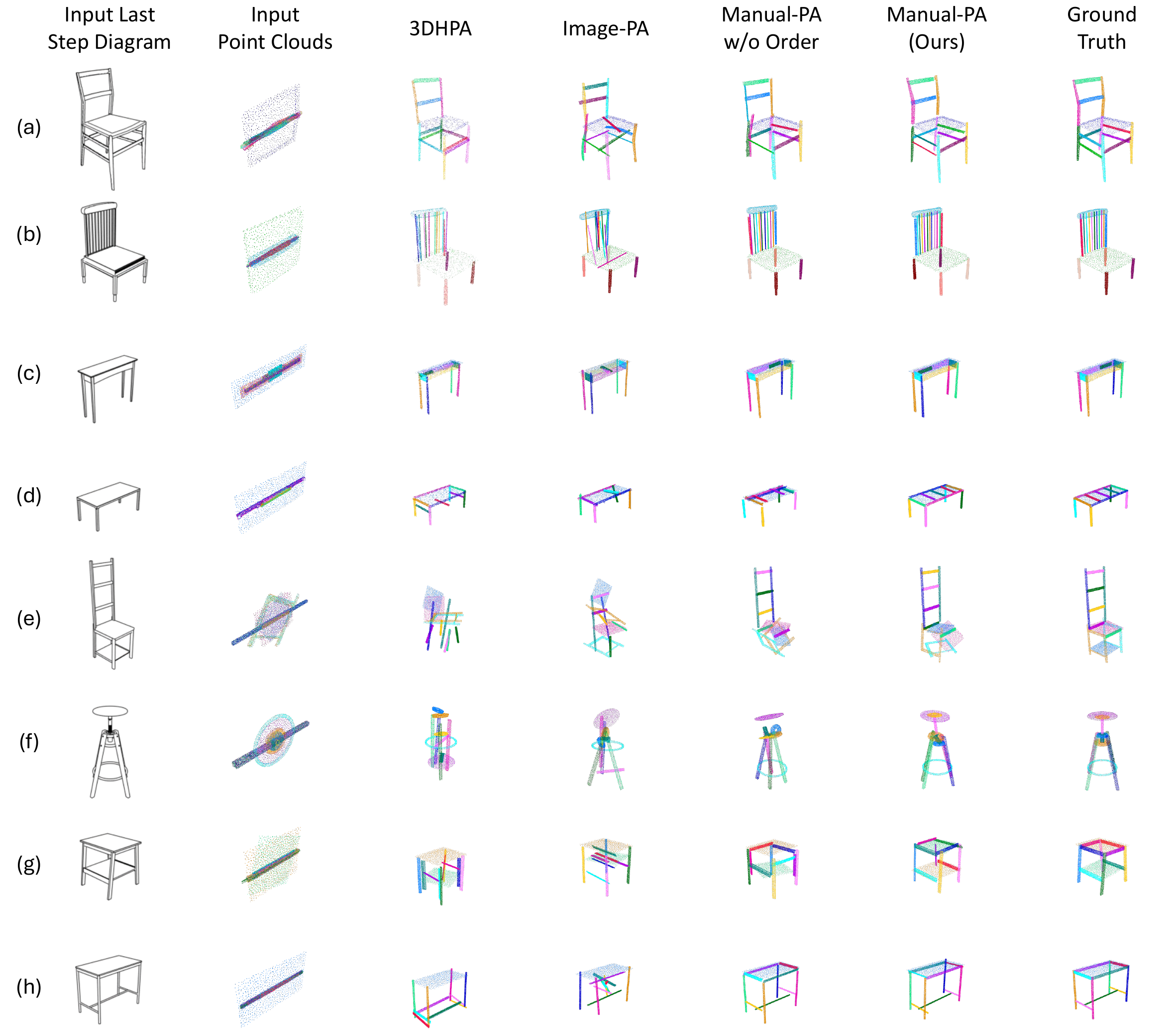}
  \caption{Qualitative comparison of various 3D part assembly methods. Eight examples are shown: chair (a), (b) and table (c), (d) from the PartNet dataset, and chair (e), (f) and table (g), (h) from the IKEA-Manual dataset.}
  \label{fig:comparision}
  \vspace{2cm}
\end{figure*}

\subsection{More Comparisons}

\clearpage

\begin{figure*}[b]
  \centering
  \includegraphics[width=\linewidth]{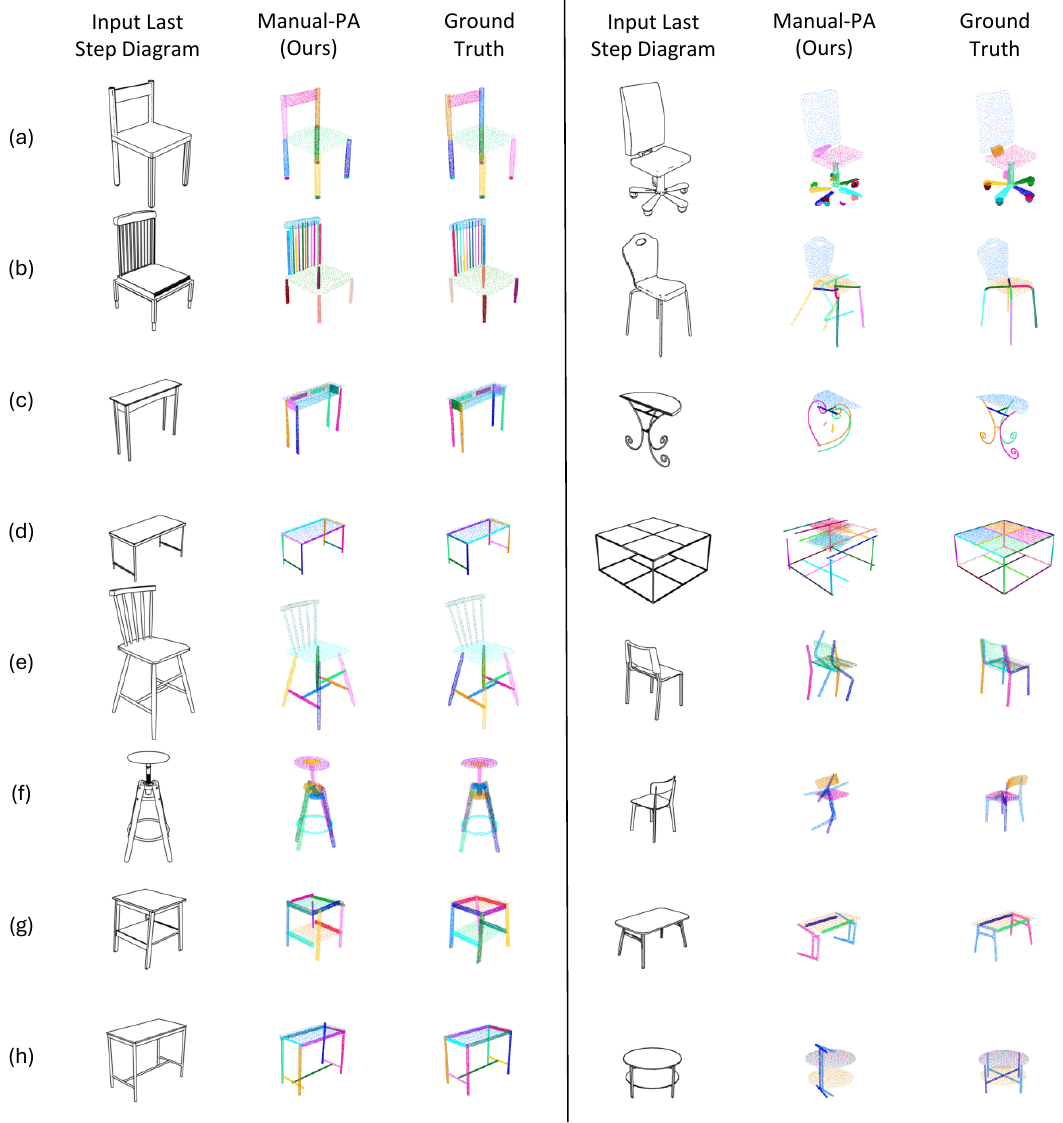}
  \caption{Results visualization of our method Manual-PA. The left column showcases examples where the method performs well, while the right column illustrates cases with less satisfactory outcomes. Eight examples are included: chairs (a), (b) and tables (c), (d) from the PartNet dataset, and chairs (e), (f) and tables (g), (h) from the IKEA-Manual dataset..}
  \label{fig:more_visualization}
  \vspace{1cm}
\end{figure*}

\subsection{More Visualizations}

\clearpage

\begin{figure*}[b]
  \centering
  \includegraphics[width=\linewidth]{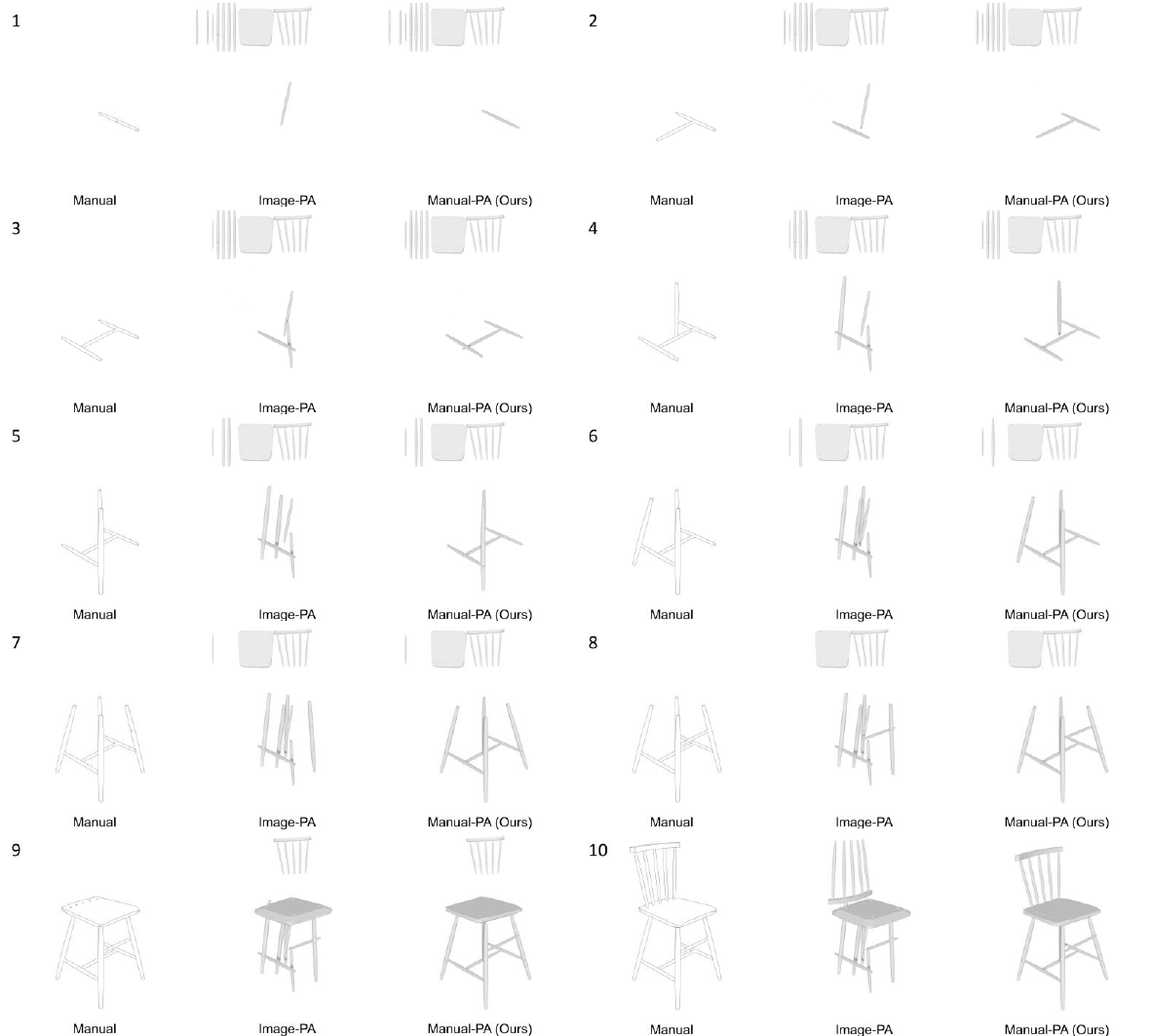}
  \caption{Snapshots from the demonstration video of the assembly process. For each frame, the leftmost column displays the step diagram from the manual, the middle column shows the assembly result of Image-PA, and the rightmost column presents the assembly result of our Manual-PA method. End frames of each step are selected for illustration. The video is provided as a part of supplementary material.}
  \label{fig:video}
  \vspace{1.75cm}
\end{figure*}

\subsection{Demonstration Video}


\clearpage

{
    \small
    \bibliographystyle{ieeenat_fullname}
    \bibliography{main}
}